\documentclass{article}
\usepackage{spconf}
\usepackage{algpseudocode}
\usepackage{algorithm}
\usepackage{amssymb, amsmath}
\usepackage{enumitem}
\usepackage{rotating}
\usepackage{booktabs}
\usepackage{multirow}
\usepackage{graphicx, epstopdf}
\usepackage{wrapfig}
\usepackage{caption}
\usepackage{subcaption}
\usepackage{adjustbox}
\graphicspath{{./figures/}}

\title{Study of Manifold Geometry using Multiscale 
Non-Negative Kernel Graphs}
\name{Carlos Hurtado$^\star$, Sarath Shekkizhar$^\dagger$, Javier Ruiz-Hidalgo$^\star$, Antonio Ortega$^\dagger$}
\address{$^\star$Universitat Polit\`ecnica de Catalunya, Barcelona, Spain\\
$^\dagger$University of Southern California, Los Angeles, USA
\thanks{Our work was supported in part by DARPA grant (FA8750-19-2-1005) in the Learning with Less Labels (LwLL) program.}}

\begin{document}
\ninept
\maketitle
\begin{abstract}

Modern machine learning systems are increasingly trained on large amounts of data embedded in high-dimensional spaces. Often this is done without analyzing the structure of the dataset.  
In this work, we propose a framework to study the geometric structure of the data. 
We make use of our recently introduced non-negative kernel (NNK) regression graphs to estimate the point density, intrinsic dimension, and linearity of the data manifold (curvature). We further generalize the graph construction and geometric estimation to multiple scales by iteratively merging neighborhoods in the input data. 
Our experiments demonstrate the effectiveness of our proposed approach over other baselines in estimating the local geometry of the data manifolds on synthetic and real datasets.

\end{abstract}
\section{Introduction}
\label{sec:intro}


The geometry of a dataset can be summarized using properties such as point density, curvature, and intrinsic dimensionality~(ID).
The ID of a dataset refers to the minimum number of parameters required for its characterization while maintaining its structure~\cite{bennett1969intrinsic}. 
Approaches for ID estimation~\cite{campadelli2015intrinsic} often rely on the construction 
of similarity-based graphs such as K-nearest neighbor (KNN) or $\varepsilon$-neighborhood graphs ($\varepsilon$-graphs).
However, the choice of these ``neighborhood parameters'' (K/$\varepsilon$) is generally ad hoc, which can severely affect the estimation of ID and other geometric properties of the data~\cite{verveer1995evaluation,kegl2002intrinsic,levina2004maximum}. 
Furthermore, these similarity-based graph methods define the scale at which the geometry is estimated through the choice of the same neighborhood parameters (K/$\varepsilon$). 
As a consequence, the only way to analyze the data at different scales is by increasing the number of neighbors connected to a given query point. 

In this paper, 
we propose new local methods for studying the geometrical properties of manifolds, using novel metrics we have developed from local data neighborhoods defined with the non-negative kernel (NNK) regression graphs~\cite{shekkizhar2019graph, shekkizhar2020graph}. 
An NNK graph is built by first selecting an initial neighborhood, e.g., a KNN graph, and then using optimization to eliminate connections to geometrically redundant neighbors. 
NNK graphs are more robust to the initial neighborhood definition (e.g., KNN graphs with different K choices can lead to the same NNK graph). More importantly, the number of NNK neighbors is explicitly dependent on  
the local geometry of the data and results in a local polytope around any query point.

We propose new metrics derived from NNK graphs to gain insights into three aspects of the geometry of a manifold. 
%
First, we study local manifold properties directly derived from NNK neighbors. The NNK optimization implies that the number of points selected in a neighborhood and the size of the local polytope depend on the local geometry of the data. 
This is in contrast to the measures obtained from a KNN or $\varepsilon$-graph where the number of neighbors or the diameter will depend only on the threshold parameters.

Second, we propose local linear subspace estimation through low-rank approximation of similarity-based graphs via principal component analysis (PCA) on the features of the points in each neighborhood. These lower-dimensional projections are associated with the tangent plane to the manifold's surface, and their dimension has been used~\cite{fukunaga1971algorithm,bruske1998intrinsic,little2011multiscale,kaslovsky2011optimal} to estimate ID.
For nonlinear manifolds, the low-rank approximations of similarity-based graphs have been shown~\cite{verveer1995evaluation,kegl2002intrinsic,levina2004maximum} to depend heavily on the definition of the local neighborhoods, and therefore, on the choice of the threshold parameter (K/$\varepsilon$).
In the NNK neighborhood optimization, only one point in each direction will be selected, and, while only locally relevant directions will be chosen, stronger directions will not be reinforced. This way, while NNK is more robust to nonlinearities in the data representation space, KNN projections will be more robust when there is linearity. This way, the change in KNN projections as a function of scale can be useful in assessing linearity, while NNK low-rank approximations will provide more reliable estimates for the local tangent plane to the manifold.

Finally, we propose a geometric analysis at multiple scales. Different approaches have been proposed to analyze manifolds at multiple scales. \cite{little2011multiscale,kaslovsky2011optimal} build on the technique of applying PCA locally by taking a multiscale approach in the KNN graph construction. This approach, however, relies on choosing an appropriate range of values for $K \geq \text{ID}$, where the value is small enough that the manifold is linear and large enough to mitigate noise in the data. Thus, \cite{little2011multiscale,kaslovsky2011optimal} are highly sensitive to the density and distribution of points in the manifold. An alternative approach followed by TwoNN~\cite{facco2017estimating} is to work on smaller subsets of the initial dataset, generated by random sampling of data points. 
The estimates are then aggregated, under the assumption that the errors that arise from the sampling will average to zero for a large enough number of subsets. In practice, ID estimates based on random sampling have high variance on sparse manifolds and do not account for changes in the local manifold structure.

We propose an alternative approach to dataset sampling in which the points in the dataset are merged iteratively based on the neighborhood defined at the current scale. This step is repeated until a dataset of the desired size is obtained. Subsets with different geometrical properties can be achieved based on the choice of similarity metric and neighborhood definition. For example, when using a Euclidean distance-based KNN similarity graph, the closest points in space will be selected and denser areas will be merged faster. On the contrary, when using the distance-based similarity but with NNK similarity graphs we can preserve the geometry of the initial data and maintain areas of different density in the resulting sampled datasets.

In summary, we propose a framework to study the local geometry of data using the properties of NNK graphs. We demonstrate via experiments: (i) ID estimation using NNK is in line with the state-of-the-art methods, (ii) linearity of data manifolds using KNN and NNK graphs, and (iii) the impact of neighborhood choice in merging examples for scale. Practical applications of some of the metrics (at one scale) presented here are studied in the context of transfer performance of pre-trained neural networks in \cite{cosentino2022geometry}.



\section{Non-Negative Kernel (NNK) regression graphs}
\label{sec:pre}

A positive definite kernel $k(\boldsymbol{x_i},\boldsymbol{x_j})$ corresponds to a transformation of points in $\mathbb{R}^d$ to points in a Hilbert space $\mathcal{H}$, such that similarities can be interpreted as dot products in this transformed space (generally referred to as \textit{Kernel Trick}). This way, $k(\boldsymbol{x_i},\boldsymbol{x_j}) = \phi_i^T \phi_j$, where $\phi: \mathbb{R}^d \rightarrow \mathcal{H}$ and $\phi_i$ represents the transformed observation $\boldsymbol{x_i}$. A popular kernel based on the distance between points that has this property is the Gaussian kernel,
\begin{equation}\label{RBF}
    k(\boldsymbol{x_i},\boldsymbol{x_j})=\exp{\left( -\frac{||\boldsymbol{x_i}-\boldsymbol{x_j}||^2}{2\sigma^2}\right)},
\end{equation}
where $\sigma$ corresponds to the bandwidth parameter of the kernel.

A KNN (or $\varepsilon$-graph) can be constructed by choosing the $K$ largest inner products $\phi_i^T \phi_j$ (or those above a threshold $\varepsilon$). Therefore, these approaches are analogous to a sparse approximation of $\phi_i$ using a thresholding approach.

In contrast, an NNK~\cite{shekkizhar2020graph} graph corresponds to an improved strategy for representation using basis pursuit. Starting from an initial KNN or $\varepsilon$-neighborhood $\mathcal{S}$, the NNK neighborhood at each node is obtained by solving
\begin{equation}\label{NNK}
    \boldsymbol{\theta}_{\mathcal{S}}=\min _{\boldsymbol{\theta}: \boldsymbol{\theta} \geq 0}\left\|\phi_i-\boldsymbol{\Phi}_{\mathcal{S}} \boldsymbol{\theta}\right\|_2^2,
\end{equation}
where $\boldsymbol{\theta}_{\mathcal{S}}$ corresponds to the weights of neighbors~($\boldsymbol{\Phi}_{\mathcal{S}}$) used to approximate $\phi_i$. Using the \textit{Kernel Trick}, the objective in \eqref{NNK} can be rewritten  as:
\begin{equation}
    \boldsymbol{\theta}_{\mathcal{S}}=\underset{\boldsymbol{\theta}: \boldsymbol{\theta} \geq 0}{\operatorname{argmin}} \frac{1}{2} \boldsymbol{\theta}^T \boldsymbol{K}_{\mathcal{S}, \mathcal{S}} \boldsymbol{\theta}-\boldsymbol{K}_{\mathcal{S}, i}^T \boldsymbol{\theta},
\end{equation}
where $\boldsymbol{K}_{i, j}=k\left(\boldsymbol{x}_i, \boldsymbol{x}_j\right)$. Thus, the $i$-th row of the graph adjacency matrix $\boldsymbol{W}$ is given by $\boldsymbol{W}_{i,\mathcal{S}}=\boldsymbol{\theta}_\mathcal{S}$ and $\boldsymbol{W}_{i,\mathcal{S}^c}=0$.

NNK performs a selection similar to the orthogonal step in orthogonal matching pursuits \cite{tropp2007signal} which makes NNK robust to the choice of sparsity parameters in the initialization (i.e., $K$ in KNN). 
Additionally, the resulting graph has a geometric interpretation where each edge in an NNK graph corresponds to a hyperplane with normal in the edge direction, points beyond which are not connected (edge weight zero) \cite{shekkizhar2019graph, shekkizhar2020graph}.

NNK has been shown to perform well in several machine learning tasks \cite{shekkizhar2021revisiting}, image representation~\cite{shekkizhar2020efficient}, and generalization estimation in neural networks~\cite{shekkizhar2021model}. Furthermore, NNK has also been used to understand convolutional neural networks (CNN) channel redundancy~\cite{bonet2022channel} and to propose an early stopping criterion for them~\cite{bonet2021channel}. Graph properties (not necessarily based on NNK graphs) have been also proposed for the understanding and interpretation of deep neural network performance \cite{gripon2018inside}, latent space geometry \cite{lassance2020deep,lassance2021laplacian} and to improve model robustness \cite{lassance2021representing}. The specific contribution of this work is to explore the effectiveness of NNK graphs in providing insights into the local geometry of the data, which can be useful in understanding the properties and structure of the whole dataset.
The metrics we propose are not limited to analyzing features in deep neural networks and can be on any dataset embedded in some space.

\begin{figure}[h!]
    \centering 
    \includegraphics[width=\linewidth]{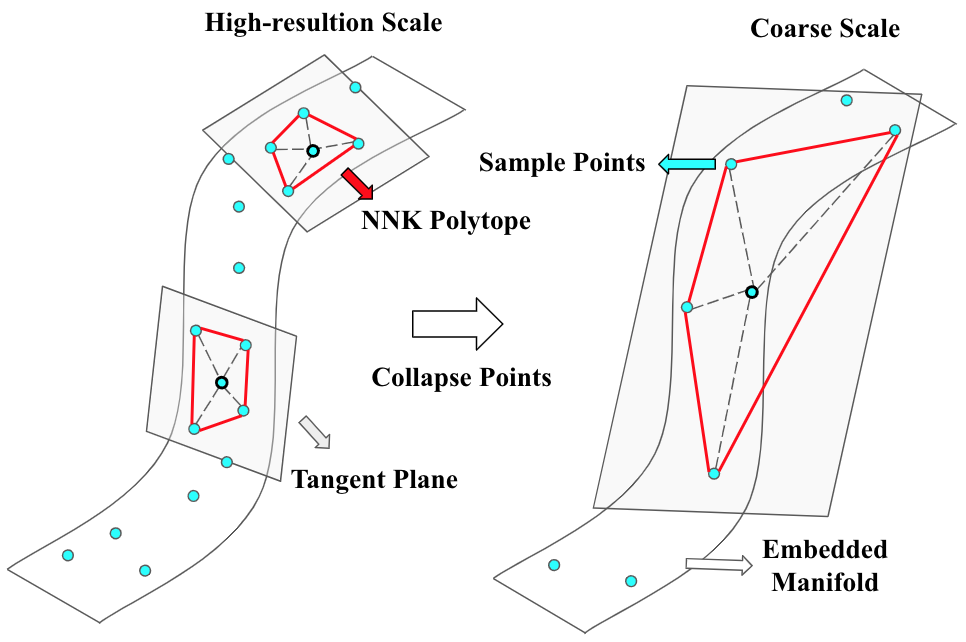}
    \caption{Our proposed approach to the geometric analysis of data using multi-scale NNK graphs. We assume observed data belongs to a manifold embedded in a higher dimensional space. Properties derived from NNK graphs allow us to capture the local geometry of the data. Changes in the properties of NNK graphs at multiple scales reflect changes in the manifold geometry.}
    \label{fig:manifold}
\end{figure}

\section{Multi-scale Analysis of NNK Graphs}
\label{sec:multi}

\begin{figure*}[h!]
    \centering
    \begin{subfigure}{0.45\linewidth}    \includegraphics[width=\linewidth]{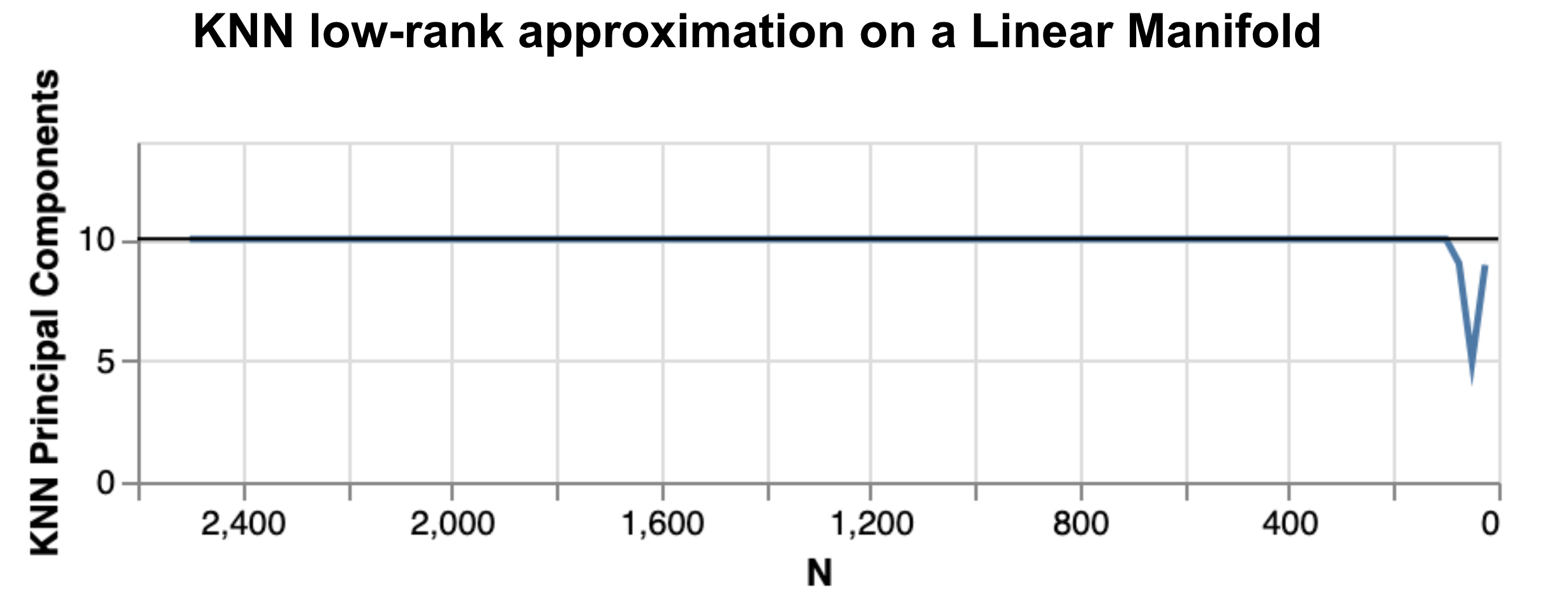}
    \end{subfigure}
    \begin{subfigure}{0.45\linewidth}
    \includegraphics[width=\linewidth]{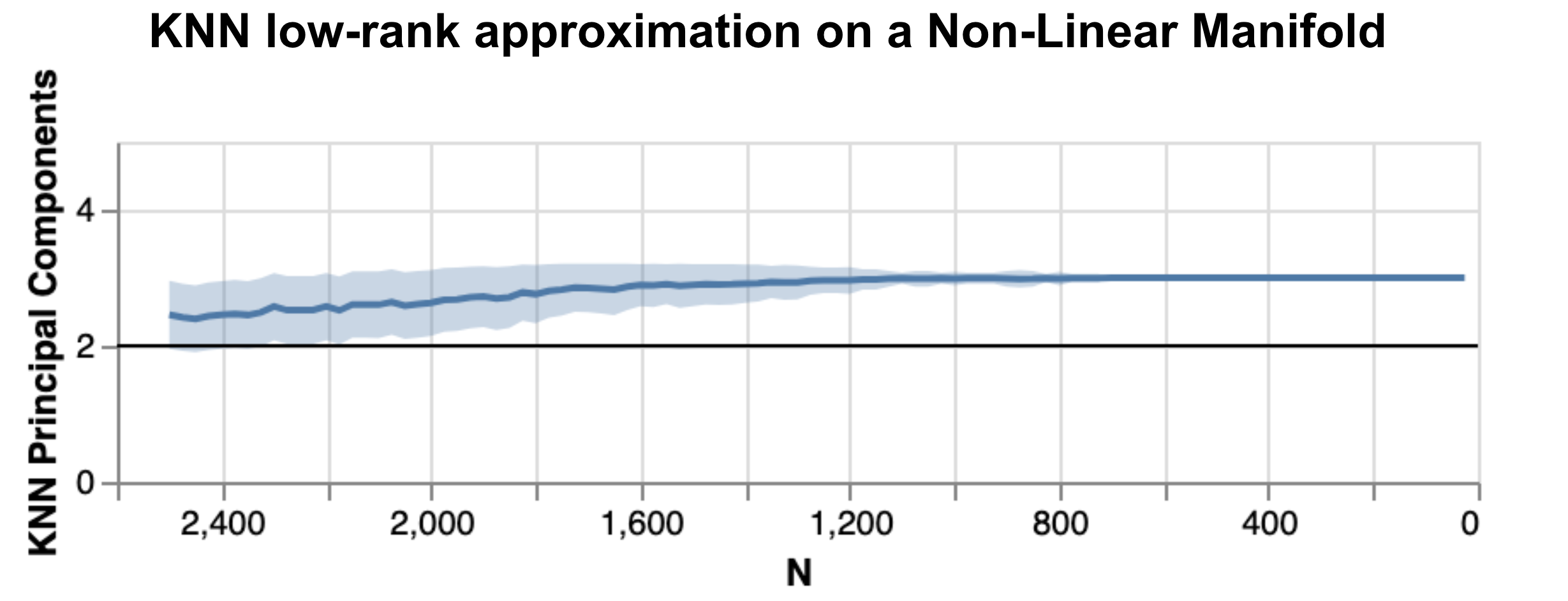}
    \end{subfigure}
    \begin{subfigure}{0.45\linewidth}    \includegraphics[width=\linewidth]{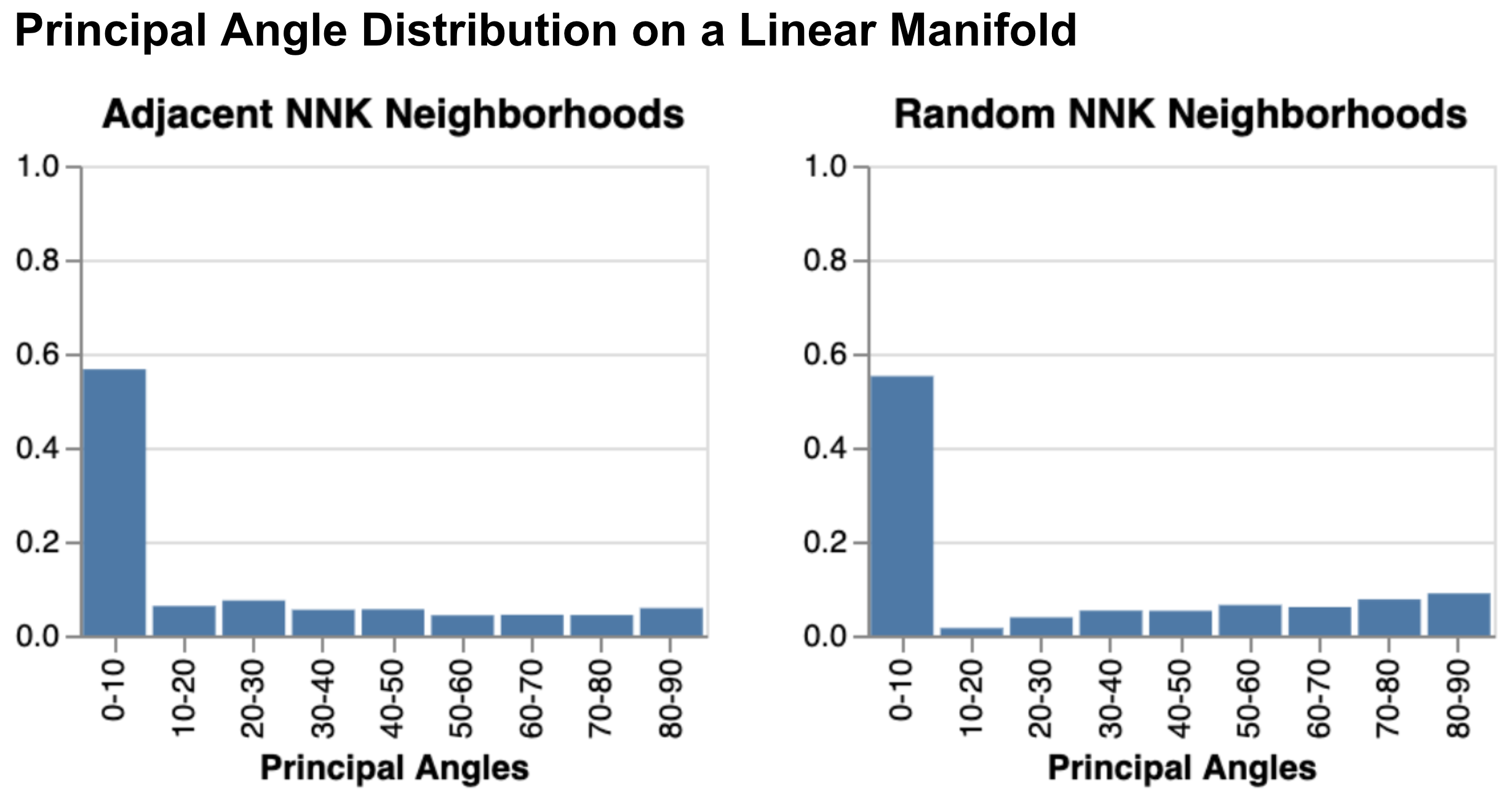}
    \end{subfigure}
    \begin{subfigure}{0.45\linewidth}
    \includegraphics[width=\linewidth]{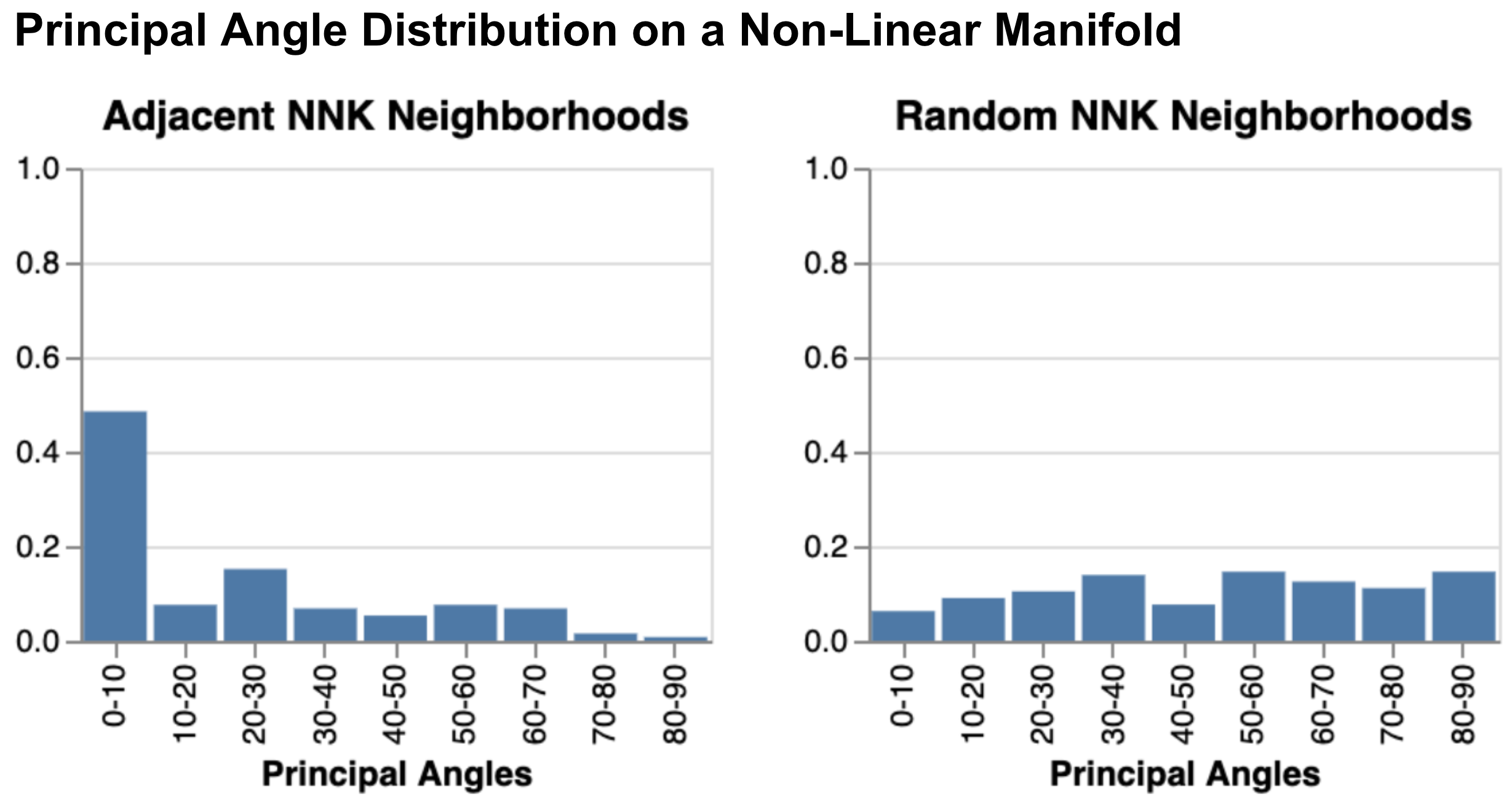}
    \end{subfigure}
    \caption{\textbf{Average number of principal components of the low-rank approximation of KNN graphs} as a function of point merging, together with a \textbf{black line indicating the datasets' ID}, and distribution of the \textbf{principal angles between NNK low-rank approximation subspaces} on a linear manifold and a non-linear manifold. We merge points using Euclidean distance, which, given that the manifolds are uniformly sampled, will result in points being merged at the same rate throughout the manifolds. We show the angles between pairs of adjacent NNK neighborhoods (i.e., one of the center nodes is a neighbor of the other) and the angles for random pairs of NNK neighborhoods. On a linear manifold, the chosen principal components of a KNN graph will be equal to the manifold's ID, and any pair of neighborhoods will have angles close to 0. On a nonlinear manifold, the principal components of a KNN graph will over-estimate the ID, and close neighborhoods have angles close to zero, while random neighborhoods are more dissimilar}
    \label{fig:synthetic}
\end{figure*}

\subsection{Local NNK neighborhood data}

For the Gaussian kernel, the local geometry of the NNK graph for a given node is a convex polytope around the node. 
Given a sufficiently large number of initial neighbors, the local connectivity of an NNK graph will be a function of the local dimension of the manifold, as depicted in Fig.~\ref{fig:manifold}. We can derive a set of geometrical properties from an NNK graph, and by comparing these properties at different points in a manifold we can gain insight into its geometry.

The number of neighbors in an NNK graph can be insightful, but it can vary locally based on (i) the distribution of the points sampled from a manifold and (ii) the location of the points relative to the geometry of the manifold (e.g., on the edges vs.~the middle of the manifold). 
We can obtain information on the local geometry of the manifold by analyzing other properties of an NNK graph.

\noindent The \textbf{diameter of an NNK polytope} is defined as the maximum distance between points in the NNK neighborhood around datapoint $k$:
\begin{equation}
    d_k = \max\limits_{i,j\in \mathcal{S}_k}\|\boldsymbol{x_i}-\boldsymbol{x_j}\|,
\end{equation}
where $\boldsymbol{x_i}$ and $\boldsymbol{x_j}$ are the features of nodes $i$ and $j$ in the NNK neighborhood of  $k$, $\mathcal{S}_k$. Given that NNK will select the nearest point along each direction in space, we can assess the local point density of a manifold using the diameter $d_k$ of polytopes around different points.

\subsection{Linear subspace estimation from an NNK graph}

An existing approach to ID estimation consists of performing a local parametrization by finding the local tangent plane in the NNK/$\varepsilon$-graph neighborhood of a point and aggregating the estimated ID for each data neighborhood analyzed~\cite{fukunaga1971algorithm,bruske1998intrinsic,little2011multiscale,kaslovsky2011optimal}. Given an appropriate neighborhood, PCA returns the local linear tangent space to the manifold. PCA estimates are robust on linear manifolds, but on nonlinear manifolds, the task of finding an appropriate neighborhood makes these estimators unstable~\cite{verveer1995evaluation,kegl2002intrinsic,levina2004maximum}.
We can also obtain a low-rank approximation from the NNK neighborhood vector subspace, such that the \textbf{number of relevant principal components} would be a robust estimate of the ID of the manifold.
The sparsity enforced by the NNK selection makes NNK graphs less sensitive to the instability of the neighborhood in the initial KNN graph.
To find the number of significant eigenvalues, we introduce a threshold on the size of the eigenvalues as proposed in~\cite{fukunaga1971algorithm}, namely
\begin{equation}
\label{eq:eigen}
    \lambda_j \geq \frac{1}{10}\cdot \lambda_{max}.
\end{equation}
%
In addition, we use the change with scale in the size of the low-rank approximations of KNN graphs to validate our observations on the curvature of a manifold. While the KNN graph's projection will not change on flat manifolds, it will be unstable and overestimate the flatness on highly curved manifolds.

By comparing the \textbf{principal angles between the low-rank approximation of NNK neighborhoods} we can better understand the geometry of a manifold.
Principal angles~\cite{jordan1875essai} refer to the generalization of the concept of angles between lines in the plane to any arbitrary dimension. This way, on a flat manifold the distribution of the angles will be similar in neighborhoods in different positions of the manifold, and many of these principal angles will be close to zero. 
In contrast, on a highly curved manifold, the distribution of the angles between NNK subspaces will change at different regions in the manifold, and the angles will be larger. Moreover, on locally smooth manifolds, we would expect the majority of the angles between the low-rank approximation of adjacent NNK neighborhoods (i.e., neighborhoods of points where one is in the NNK neighborhood of the other) to be close to zero.

\subsection{NNK graph construction at multiple scales}

To study the manifold at multiple scales, we must increase the size of the NNK neighborhood. Intuitively, we would do so by adjusting the hyperparameters of the NNK algorithm to observe points that are further away. 
However, by definition, NNK will choose a single neighbor in each relevant direction, therefore increasing the sparsity parameter $K$ in the initial KNN graph is unlikely to affect the size of the resulting NNK neighborhood after optimization. 
On the other hand, increasing the bandwidth $\sigma$ of the Gaussian kernel will make 
points that are very close to each other collapse (similarity value of $1$), in which case the NNK optimization will only select one point for the neighborhood.
To overcome this issue, we propose an approach that simultaneously increases $\sigma$ while merging the closest data points in the manifold.
Thus, we change the scale of the analysis by subsampling the manifold in a controlled way, increasing the distances between points in the process, while also increasing $\sigma$ to allow farther away points to be connected. 

We would expect a linear manifold to have similar properties at different scales. However, merging on a highly curved manifold would result in connecting points that were initially in different local neighborhoods, thus changing the shape of the manifold and leading to changes on the NNK graphs at coarser scales (see Fig.~\ref{fig:manifold}). 
Thus, studying our proposed properties at different scales can lead to a better understanding of the shape of the manifold.

To achieve a sparser representation we iteratively merge the two closest points according to some selected metric. For example, we can merge the two nodes with the shortest pairwise distance (largest KNN graph weight), and the nodes with the largest NNK pairwise weight, or use alternative metrics. 
At each step, we combine the points that are closest so that we can increase the window of observation. After each merging iteration, we can recalculate the NNK graph. Since the decay parameter $\sigma$ has been defined based on the distances in the dataset, we will in turn increase the size of the NNK graph as we merge points. 
This way, we can construct larger NNK graphs and thus be able to analyze the manifold at different scales. 
This merged dataset can be achieved as described in Algorithm 1.

\begin{algorithm}[ht]
\caption{Two-closest Merging}
\textbf{Input:} $X$ features, $I$ merging steps
\begin{algorithmic}[1]
\For {iter in $I$}
\For{each node $i=1,2,..., N$}
    \State $S_i=\{\text {neighborhood of node } i\}$ 
    \State $K_{i,S}=\{\text {similarity to each neighbor} \}$
\EndFor
\State $i,j = \{ i, j : \max\limits_{i,j} K_{i,j}\}$
\State $X = X \cup \frac{X_i+X_j}{2} \setminus X_i, X_j$
\EndFor
\end{algorithmic}
\textbf{Output:} Dataset $X$ after $I$ merging steps 
\end{algorithm}
\vspace{-5mm}

\section{Experiments}

\subsection{ID estimation benchmark}

\begin{table}[h!]
    \begin{adjustbox}{width=\linewidth}
    \begin{tabular}{ccccccccccc}
    \toprule
    \textit{d} & MLE & kNNG & BPCA & Hein & CD & DANCo & MLSVD & NNK \\
    \midrule
     \textit{1}      & \textbf{1.00} & $1.07$ & $5.70$ & \textbf{1.00} & $1.14$ & \textbf{1.00} & \textbf{1.00} & \textbf{1.00}   \\
    \textit{2}      & $2.21$ & $2.03$ & $1.55$ & \textbf{2.00} & $2.19$ & \textbf{2.00} & $1.00$ & $1.00$    \\
     \textit{2}      & $1.97$ & $2.06$ & \textbf{2.00} & \textbf{2.00} & $1.98$ & \textbf{2.00} & \textbf{2.00} & \textbf{2.00}    \\
     \textit{2}      & $1.96$ & $2.09$ & \textbf{2.00} & \textbf{2.00} & $1.93$ & \textbf{2.00} & $2.35$ & \textbf{2.00}    \\
     \textit{3}      & $2.88$ & $3.03$ & \textbf{3.00} & \textbf{3.00} & $2.88$ & \textbf{3.00} & \textbf{3.00} & \textbf{3.00}    \\
     \textit{4}      & $3.83$ & $3.82$ & \textbf{4.00} & \textbf{4.00} & $3.23$ & \textbf{4.00} & $2.08$ & \textbf{4.00}    \\
     \textit{4}      & $3.95$ & $4.76$ & $4.25$ & \textbf{4.00} & $3.88$ & \textbf{4.00} & $8.00$ & \textbf{4.00}    \\
     \textit{6}      & $6.39$ & $11.24$ & $12.00$ & \textbf{5.95} & $5.91$ & $7.00$ & $12.00$ & $8.00$    \\
    \textit{10}     & $8.26$ & $10.21$ & $5.20$ & $8.90$ & $8.09$ & $9.86$ & \textbf{10.00} & \textbf{10.00}    \\
     \textit{10}     & $9.10$ & $9.98$ & $5.45$ & $9.45$ & $9.12$ & $10.09$ & \textbf{10.00} & \textbf{10.00}    \\
     \textit{3}     & $4.05$ & $4.32$ & $4.00$ & \textbf{3.00} & $3.37$ & $4.00$ & 1.00 & \textbf{3.00}    \\
     \textit{8-11}      & \textbf{10.29} & \textbf{9.58} & \textbf{11.00} & \textbf{8.00} & 6.96 & \textbf{9.98} & 1.00 & \textbf{10.00}    \\
    \bottomrule
    \end{tabular}
    \end{adjustbox}
    \caption{ID estimation on synthetic datasets~\cite{campadelli2015intrinsic} (first 10 rows), Isomap~\cite{tenenbaum2000global}, and MNIST~\cite{lecun1998mnist} (last 2 rows), using $7$ state-of-the-art methods and our proposed NNK approach.}
    \label{tab:benchmark}
\end{table}

We estimate ID from the number of principal components selected from the NNK graph following equation (\ref{eq:eigen}). We compare our proposed method for ID estimation (NNK), with state-of-the-art methods for each of the categories defined in an ID estimator literature review and benchmark proposal~\cite{campadelli2015intrinsic}. We include BPCA~\cite{bishop1998bayesian} and MLSVD~\cite{little2017multiscale}, which are projective estimators; the kNNG~\cite{costa2005estimating} graph-based estimator; CD~\cite{grassberger2004measuring} and Hein~\cite{hein2005intrinsic} as examples of topological fractal estimators; and MLE~\cite{levina2004maximum} and DANCo~\cite{ceruti2012danco}, topological nearest neighbor-based estimators.
We show the results on a series of synthetic datasets~\cite{hein2005intrinsic} generated by uniformly drawing samples from manifolds of known ID that are embedded linearly or nonlinearly on higher dimensional spaces. We also show results for the Isomap~\cite{tenenbaum2000global} faces and MNIST~\cite{lecun1998mnist} datasets.

Table~\ref{tab:benchmark} shows that the estimate of ID derived from NNK graphs achieves performance in line with that of other estimators in the literature. While a more complete benchmark would help assess the capabilities and limitations of our NNK estimates, our results show that meaningful information about the local geometry of a data manifold can be derived from NNK neighborhoods.

\subsection{Synthetic manifold analysis}

We use two of the metrics we previously described to assess the linearity of two synthetic data manifolds. In the first row, we show the size of the low-rank approximations of KNN neighborhoods as a function of the points in the dataset. Below, we plot the distribution of the principal angles between the low-rank approximation of NNK neighborhoods. Fig.~\ref{fig:synthetic} shows the distributions obtained by comparing adjacent and random NNK neighborhoods on a linear (left) and nonlinear (right) manifold. For the linear manifold, the distribution of both adjacent and random pairs of neighborhoods are almost the same, since the geometry of a neighborhood in a linear manifold will be similar throughout the manifold. 
In contrast, on the nonlinear manifold, there is a difference in the distribution of the angles.

\begin{figure}[H]
    \centering
    \begin{subfigure}{0.49\linewidth}
    \includegraphics[width=\linewidth]{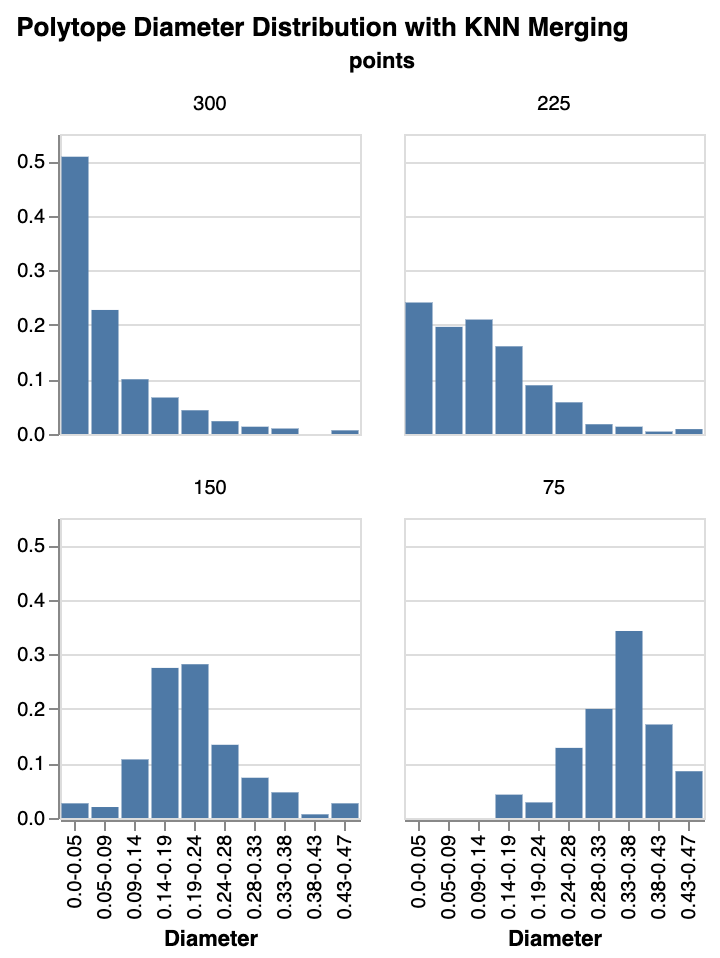}
    \end{subfigure}
    \begin{subfigure}{0.49\linewidth}
    \includegraphics[width=\linewidth]{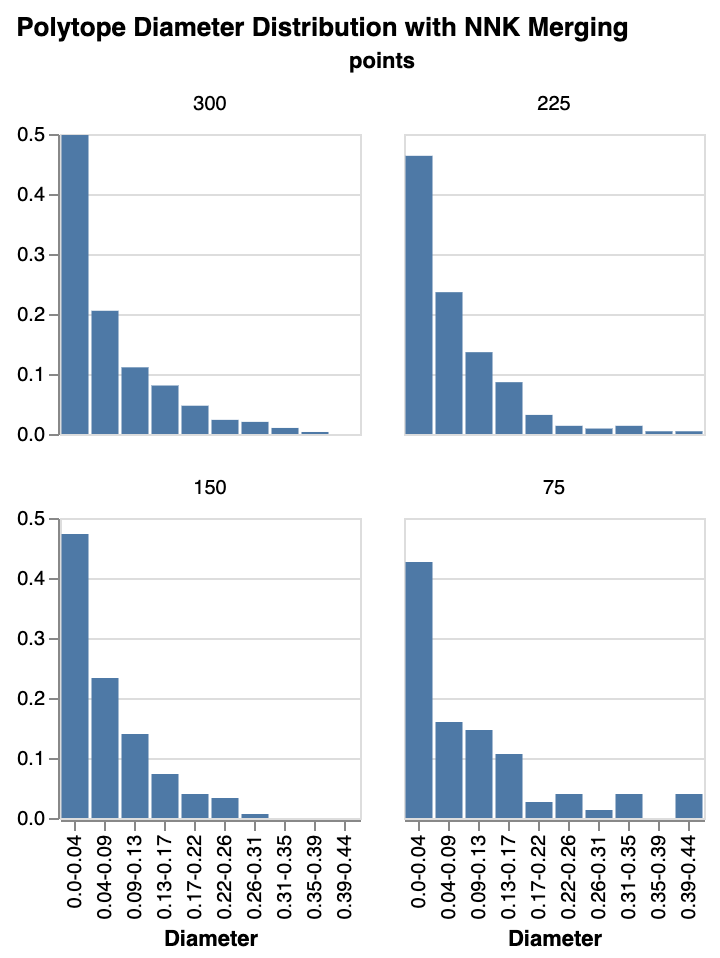}
    \end{subfigure}
    \caption{Distribution of the diameter of NNK graph polytopes constructed on the \textit{control} dataset while merging with KNN or NNK similarity. When merging based on KNN similarity, we observe the polytope diameter distribution shift to larger diameters. When using NNK similarity in the selection of the points to merge, the distribution of the polytope sizes remains the same after merging.}
    \label{fig:merge}
\end{figure}

\subsection{Choice of neighborhoods for merging}

From the distribution of the diameter of the NNK polytopes for the same dataset, in Fig.~\ref{fig:merge}, we observe that when using KNN similarity for merging, the size of the polytopes increases, such that the size distribution shifts to larger polytopes while also growing in size. 
In contrast, when NNK similarity is used for merging, the distribution of the polytope diameters is preserved as their sizes grow overall. This shows that NNK merging is better if the goal is to preserve the differences in point densities in the multiscale analysis.

\section{Conclusion}

We present a framework based on the geometrical properties of NNK graphs to gain insight into the shape of data manifolds in terms of their intrinsic dimension, curvature, and point density. The proposed metrics are the dimension of the low-rank approximation of the KNN and NNK graphs, the diameter of NNK graphs, and the principal angles between the low-rank approximations of NNK graphs. Moreover, we compare these metrics at multiple scales, which we can do by using our proposed point merging algorithm. Experiments show that we can accurately estimate ID on popular benchmark manifolds. Furthermore, we have shown the effectiveness of NNK properties in characterizing data manifolds.
\clearpage
\bibliographystyle{IEEEbib}
\bibliography{refs}


\vfill\pagebreak


\end{document}